\documentclass[letterpaper, 10 pt, conference]{ieeeconf}  

\IEEEoverridecommandlockouts                              

\usepackage{url}
\usepackage{array}
\usepackage{graphicx}
\usepackage{amsmath}
\usepackage{amssymb}
\usepackage{algorithm}
\usepackage[noend]{algpseudocode}
\usepackage[percent]{overpic}
\usepackage{etoolbox}
\usepackage{array,multirow}
\usepackage{blindtext}
\usepackage{hyperref}
\usepackage{color}
\usepackage{tabularx}
\usepackage{booktabs}
\usepackage{arydshln}

\title{\LARGE \bf
Enabling End-Users to Deploy Flexible \\ Human-Robot Teams to Factories of the Future
}

\author{Dominik Riedelbauch$^{1}$, Johannes Hartwig$^{1}$ and Dominik Henrich$^{1}$
	\thanks{
	$^1$ The authors are with the Chair for Applied Computer \mbox{Science III} (Robotics and Embedded Systems), University of Bayreuth, D-95440 Bayreuth, Germany;
	{\tt\scriptsize \{dominik.riedelbauch | johannes.hartwig | dominik.henrich\}@uni-bayreuth.de};}%
	\thanks{$^2$ This project has partly been supported by the Deutsche Forschungsgemeinschaft (DFG) under grant agreement He2696/20 FlexCobot.}%
}

\begin{document}

\maketitle
\thispagestyle{empty}
\pagestyle{empty}
\bstctlcite{IEEEexample:BSTcontrol}

\begin{abstract}
	Human-Robot Teams offer the flexibility needed for partial automation in small and medium-sized enterprises (SMEs). They will thus be an integral part of Factories of the Future. Our research targets a particularly flexible teaming mode, where agents share tasks dynamically. Such approaches require cognitive robots with reasoning and sensing capabilities. This results in hardware maintenance demands in terms of sensor calibration. In contrast to intuitive end-user programming, system setup and maintenance are rarely addressed in literature on robot application in SMEs. In this paper, we describe a prototype software toolchain that covers the initial setup, task modelling, and online operation of human-robot teams. We further show, that end-users can setup the system quickly and operate the whole toolchain effortlessly. All in all, this work aims to reduce the concern, that deploying human-robot teams comes with high costs for external expertise.
\end{abstract}

\section{BACKGROUND AND RELATED WORK}\label{section:background_and_related_work}
Robots have started to emerge from potentially hazardous tools that need to be locked behind fences to teammates working hand in hand with humans. Fence-less lightweight robots suit the needs of SMEs, where flexibility to produce varying products with small lot sizes is a key \mbox{requirement \cite{Perzylo19}}. A major part of these enterprises is already using lightweight robots, or planning to do so within few years \cite{Kildal18}. However, humans and robots merely coexist in recent applications rather than forming symbiotic teams that share work and utilize individual strengths of humans and robots \cite{Bender16}. The \textsc{FlexCobot}$^{2}$ project seeks to close this gap.

In contrast to numerous works on static human-robot task sharing (e.g. \cite{Michalos18}\cite{Johannsmeier17}), \textsc{FlexCobot} is based on \emph{dynamic task allocation} through online, iterative decisions. To this end, the approach grants decision making authority to all teammates and results in flexible human-robot teams. In particular, our notion of a \emph{flexible team} is characterized by dynamic transitions between three teaming \mbox{modes (Fig. \ref{img:flexible_team})}:  
\begin{figure}
	\centering
	\begin{overpic}[width=\linewidth]{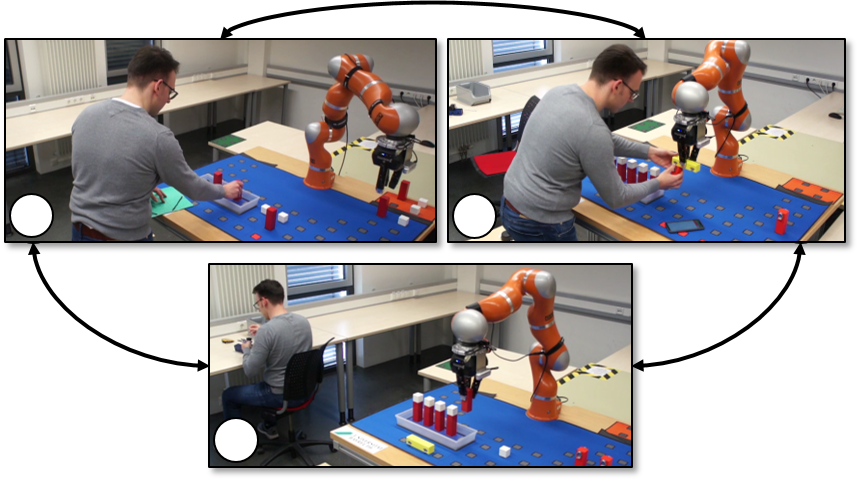}
		\put(3.2, 29.5){\small i}
		\put(54.4, 29.5){\small ii}
		\put(25.9, 3.){\small iii}
	\end{overpic}
	\caption{Flexible Human-Robot Teams can switch between cooperation (i), collaboration (ii) and coexistence (iii) dynamically.}
	\label{img:flexible_team}
\end{figure}
(i) \textbf{Cooperation}: In this mode, human and robot work efficiently by carrying out sub-tasks of the same task in parallel. (ii) \textbf{Collaboration}: In contrast to the loose coupling of cooperation, partners work on the same sub-task during collaboration. Physical contact, potentially transmitted via parts to be handled jointly, is intended. (iii) \textbf{Coexistence}: Transitioning into coexistence allows workers to flexibly handle urgent intermediate tasks (e.g. handling a delivery of goods or supplying the workstation with material) by leaving the workbench temporarily. Our approach provides the robot with sufficient autonomy to keep working in the meantime. Humans may decide to re-join into cooperation at any time. The robot is equally able to initiate mode transitions, e.g. by actively calling an absent partner for help within a collaborative sub-task. More details on this task allocation method based on skill interaction categories and human-aware world modelling are given in our previous publications \cite{Riedelbauch19} and \cite{Riedelbauch19b}. Video clips are available online\footnote{\scriptsize \url{http://robotics.uni-bayreuth.de/projects/flexcobot/}}.

Although flexibility is a key requirement, this feature is not the only concern that must be addressed when designing robot systems for Factories of the Future -- the cost factor must also be taken into account, especially with regard to the perceived lack of in-house programming expertise and expected personnel expenditure for external experts \cite{Perzylo19}\cite{Kildal18}. This problem has been approached by novel methods for intuitive end-user programming of industrial robots, which are often based on task modelling via graphical user interfaces (e.g. \cite{Steinmetz18}\cite{Paxton17}\cite{Kraft17}). These methods reduce the need for external programmers, as they enable the existing workforce to teach the system. Flexible human-robot teaming poses additional requirements on task modelling. A suitable task model must encode potential parallelism of sub-tasks for efficient cooperation. It must moreover supply robots with means to observe and understand task progress. We have previously shown, that the advantages of graphical end-user programming can be transferred to commissioning of flexibly shared tasks in spite of these increased requirements \cite{Riedelbauch18}.
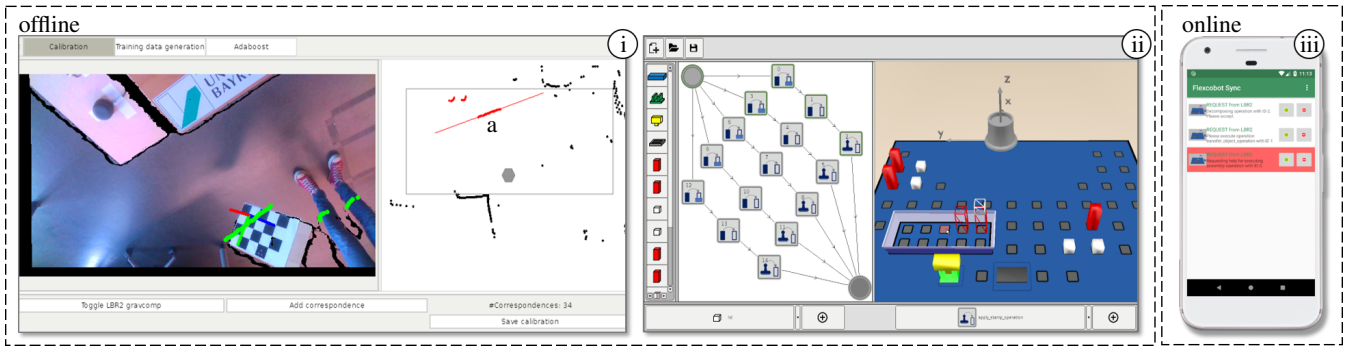
\begin{figure*}[h!]
	\centering
	\begin{overpic}[width=\textwidth]{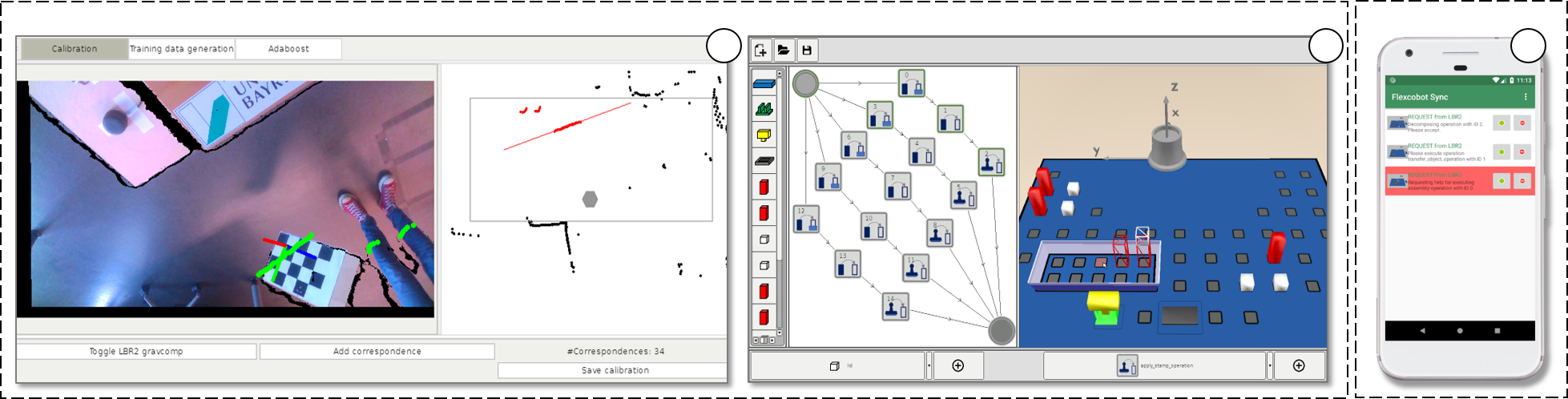}
		\put(1, 23.5){\small offline}
		\put(88, 23.5){\small online}
		\put(46, 22){\small i}
		\put(84.2, 22){\small ii}
		\put(96.81, 22){\small iii}
		\put(36, 16){\small a}
	\end{overpic}
	\caption{Our software toolchain consists of tools for calibration (i), task modelling (ii) and human-robot communication (iii) that support the offline and online phase of flexible human-robot teaming.}
	\label{img:software_toolchain}
	\vspace{-0.5cm}
\end{figure*}

The issue of understanding task progress brings us to an additional problem that must generally be dealt with when adapting robots to new tasks: Smart robot assistants do not follow a program thoughtlessly, but also use sensors to perceive and react to their environment. Usually, these sensors need to be calibrated to the robot to provide a unified coordinate frame, e.g. for reasoning. Having in mind the cost argument that motivates intuitive programming, this step should also be feasible for end-users. The importance of user-friendly ways to calibrate a RGB-D vision system has previously been stated by Paxton et al. in the context of intuitive programming \cite{Paxton17}, and by Rückert et al. \cite{Rueckert18} for the field of human-robot collaboration. A convenient one-click solution for calibrating an industrial manipulator to a multi-camera system for safe human-robot coexistence has been proposed by Werner et al. \cite{Werner18}. Overall systems including the calibration step are however not addressed in recent literature on teaming with dynamic task sharing (e.g. \cite{Darvish18}\cite{Nikolakis18}). 

Combining the aforementioned ideas and requirements, the contribution of this short paper is as follows: We will first describe our prototype software toolchain for flexible human-robot teaming in Section \ref{section:software_toolchain_for_flexible_teaming}. This toolchain integrates support of end-users throughout all stages of system operation, from the offline steps of initial calibration and intuitive task modelling to online worker support and human-robot communication. The contribution lies not primarily in the composition of these software tools -- we rather use the prototype as a basis for showing, that deploying flexible human-robot teams is feasible for non-expert users within a short timespan (\mbox{Section \ref{section:evaluation_and_results}}). To the best of our knowledge, no comparable overall system for teaming has yet been considered from this point of view in existing literature.

\section{A SOFTWARE TOOLCHAIN FOR \\ FLEXIBLE TEAMING}\label{section:software_toolchain_for_flexible_teaming}
The structure of our software toolchain is shown in \mbox{Fig. \ref{img:software_toolchain}}. The \emph{offline phase} covers the steps of calibration (i) and task modelling (iii). Based on the resulting task models, human and robot may work as a team in the \emph{online phase}. This phase is supported by a smartphone app for human-robot communication (iii). A detailed view on the individual steps will be given hereinafter. The focus will lie on the technical details of calibration, as this step complements our previous work on task modelling \cite{Riedelbauch18} and shared task execution \cite{Riedelbauch19}\cite{Riedelbauch19b}. 

\begin{figure}
	\centering
	\begin{overpic}[width=0.8\linewidth]{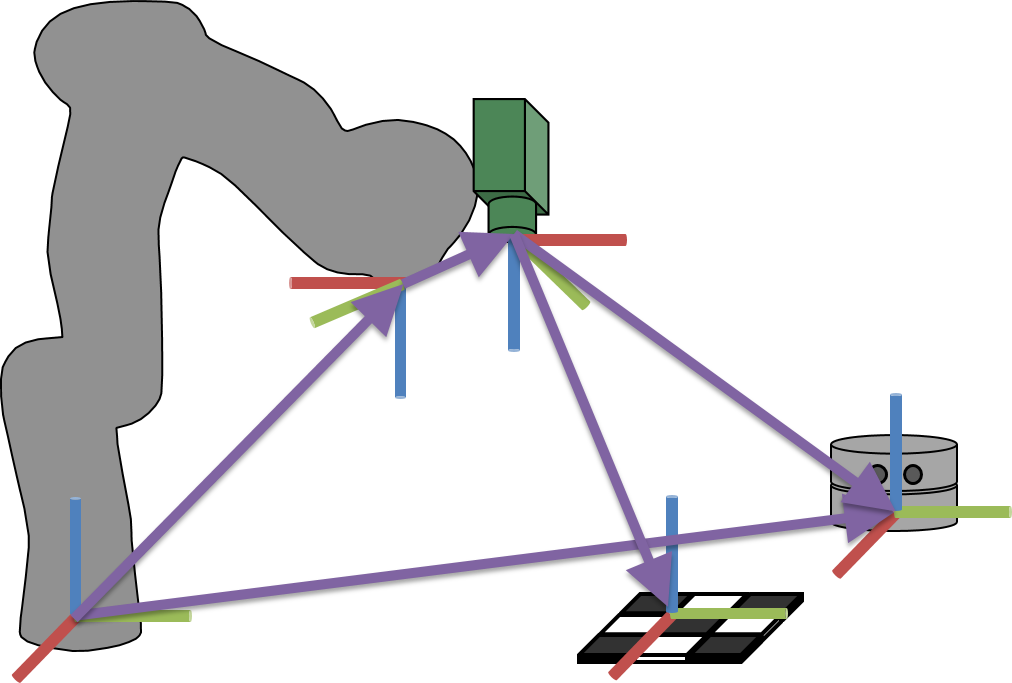}
		\put(2, 5){$W$}
		\put(25.2, 20){$B_k$}
		\put(42.2, 36.){$X$}
		\put(49, 6.2){$Y$}
		\put(53., 22){$A_k$}
		\put(63.2, 27.5){$C_k$}
	\end{overpic}
	\caption{Solving the calibration problem involves transformations between world frame and robot TCP ($B_k$), TCP and camera coordinates ($X$), camera coordinates and LIDAR data frame ($C_k$), world frame and LIDAR data frame ($Y$), and the calibration pattern pose wrt. the camera frame ($A_k$).}
	\label{img:calibration_problem}
\end{figure}
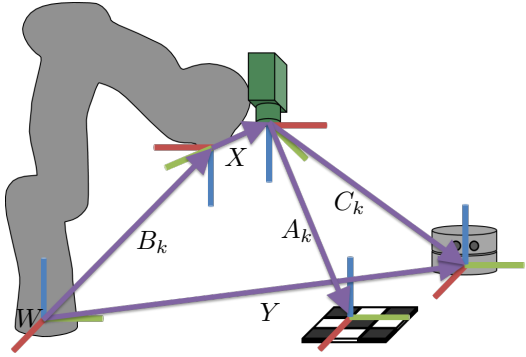

\textbf{Calibration:} The \textsc{FlexCobot} system makes decisions based on sensor data from two sources: An eye-in-hand camera mounted near the robot hand is used to maintain a symbolic world model by recognizing parts in the workspace. This avoids the occlusions that cameras in fixed positions might suffer from. Of course, humans may manipulate previously detected parts, while they are out of sight for the robot -- inspired by the process of human forgetting, the system deals with partial observability by losing trust in parts, when they enter the sphere of influence of any \mbox{human \cite{Riedelbauch19}}. To this end, humans are tracked in the data provided by a 2D LIDAR sensor near the workbench. Determining, whether humans can access certain objects in the world model requires part and human positions to be known in the same coordinate frame. This brings us to the following two-staged multi-sensor calibration problem:

In order to place parts perceived by the camera in the coordinate frame $W$ of the robot world model, we first need to know the homogenous extrinsic calibration matrix $X\in\mathbb{R}^{4\times 4}$ between tool centre point (TCP) and camera frame (cf. Fig. \ref{img:calibration_problem}). This problem of eye-in-hand calibration is well-known and commonly solved by observing a calibration pattern from $N$ different camera poses. Then, $X$ can be optimized to satisfy
\begin{align}
	A_i X B_i = A_j X B_j \Leftrightarrow A_{ij}X = XB_{ij},
\end{align}
where $A_{ij} = A_j^{-1}A_i$, $B_{ij} = B_j B_i^{-1}$, $i, j\in \{1, ..., N\}$, $i\neq j$. $B_k$ is known from robot forward kinematics. The pattern position $A_k$ in the camera frame can be directly calculated from camera images since we use a RGB-D camera. An overview of approaches to solve this optimization problem is given by Shah et al. \cite{Shah12}. We used the dual quaternion approach proposed by Daniilidis \cite{Daniilidis98} and refined the result with an iterative non-linear least squares optimization. Corresponding $A_k$ and $B_k$ can be collected using a graphical user interface that moves the robot to predefined poses upon one single mouse click, and starts the optimization afterwards. The calibration pattern is shaped to fit the robot base segment to ensure appropriate placement. 

We furthermore need to transform human positions detected by the laser range finder into the world model coordinate frame $W$. As soon as $X$ is known, the extrinsic LIDAR calibration matrix $Y\in \mathbb{R}^{4\times 4}$ between the world model and LIDAR data frame is given as
\begin{align}
	\label{eq:lidar}
	Y = C_k X B_k,
\end{align}
for some homogeneous transform $C_k \in \mathbb{R}^{4\times 4}$ between camera and LIDAR frame and corresponding robot pose $B_k$.
If we now choose a fixed robot pose $B_f$, we can reduce the problem to calculating a fixed $C_f$, i.e by deriving the extrinsic calibration parameters between LIDAR sensor and eye-in-hand camera. This problem can be solved with the method of Zhang and Pless \cite{Zhang04}. To this end, the calibration pattern needs to be placed at $M$ different positions in view of the camera, while also being sensed by the LIDAR sensor. At each position $l=1...M$, the calibration plate is seen as a set of samples $P_l$ forming a line segment in the LIDAR scan (Fig. \ref{img:software_toolchain}(i)a, red line fit). We know the normal vector $N_l$ of the plate from the RGB-D camera image. All samples $p \in P_l$ must lie on the plane defined by $N_l$ after projecting them into the camera coordinate frame. Thus,
$
		N_l \circ C_f p = \lVert N_l \rVert^2
$
must hold for all $p\in P_l$. This leaves us with $|P_l|$ equations at each of the $M$ pattern positions. $C_f$ results from optimization over the union of these equations. Using the resulting $C_f$ in Formula \ref{eq:lidar} provides the searched value of $Y$. Gathering point-plane correspondences at different pattern positions is supported by a calibration app (Fig. \ref{img:software_toolchain}(i)). The app allows users to reposition the robot so the camera and the LIDAR sensor can detect the pattern simultaneously. While moving the pattern, the app gathers correspondences, solves for $Y$ repeatedly, and projects LIDAR samples into the camera image (green in Fig. \ref{img:software_toolchain}(i)). Users may stop the calibration process, as soon as the reprojected samples are seen to cover their legs and the plate sufficiently.

\textbf{Task Modelling:} The graphical editor for shared tasks that we presented initially in \cite{Riedelbauch18} is the second software component of the toolchain. This editor is based on a robot skill framework inspired by the work of Andersen et al. \cite{Andersen14}. So called motor and perception primitives (e.g. \verb|Grasp|, \verb|Place|, \verb|Hold|, \verb|Transfer|, \verb|TriggerCamera|) are grouped into more complex skills by specifying a graph for control and data flow. We addressed the teaming aspect by demanding a well-defined graph structure, so that object-centric, observable preconditions and postconditions can be deduced from parametrized skills automatically. We moreover added \emph{communication primitives} (e.g. \verb|WaitForAcknowledge|) to synchronize the work of human and robot by communicating within a skill. Skills created by experts or system integrators (e.g. \verb|PickAndPlacePart|, \verb|MateParts| in the current prototype implementation) open modelling of more complex tasks to end-users \cite{Steinmetz18}. To this end, our editor adopts ideas from CAD-based robot programming. Parts can be instantiated by placing them within a virtual environment (Fig. \ref{img:software_toolchain}(ii), right). They are then used as parameters to skills. It is important to notice at this point, that skill parametrization is solely intended to reflect the process to be modelled. We do not assume, that end-users will take into account robot characteristics (e.g. limited reach due to kinematic properties) when specifying the position of objects on the workbench -- the resulting issues are resolved by communication during the online execution phase. Each parametrized instance of a skill is represented by a pictogram (Fig. \ref{img:software_toolchain}(ii), left) referring to the skill type. Users may establish precedence relations among skills by connecting these pictograms. The overall task modelling process results in precedence graphs as known from the assembly planning domain. Future work may thus open the toolchain to task models originating from automated assembly planning \cite{Niu03}. 

\textbf{Online Task-Sharing:} Precedence graphs created with the task editor are shared with the robot via an XML representation. Pre- and postconditions of skills help the system to estimate task progress by matching conditions against the state of detected objects in the world model. With this knowledge on progress and the aforementioned trust in world model content, our robot teammate can iteratively try to execute skills that are likely to succeed \cite{Riedelbauch19}. As stated before, skills in the task model must not necessarily be feasible for the robot, as this would require expert knowledge in the modelling step. Hence, the dynamic task allocation algorithm is moreover able to reason about feasibility by classifying skills into interaction categories \cite{Riedelbauch19b}. These categories mirror the degree of interaction that is needed to work off some skill: The robot may e.g. execute a skill all by itself, fully delegate it to the human, or enter collaboration. The latter two cases result in a need for communication. Our software toolchain provides a smartphone application for this purpose (Fig. \ref{img:software_toolchain}(iii)). This is motivated by the facts that smartphones are already ubiquitous and will certainly be at hand for workers in Smart Factories of the Future. Via this app, the robot may e.g. inform its partner about skills that it is not capable of, or ask the human to switch into the collaboration mode via short messages and visual cues.

\section{EVALUATION AND RESULTS}\label{section:evaluation_and_results}
In this section, we will first present novel results from the evaluation of the calibration app. A recapitulation of previous experiments on task modelling and execution enables us to discuss the issue of end-users operating the whole toolchain.

\textbf{Calibration:} 
Hereinafter, we investigate the hypothesis, that our software tools for calibration enable end-users to perform a complex multi-sensor calibration comfortably within a reasonable amount of time. Against the background of industrial applications in SMEs, participants with a technical background were chosen. None of them indicated having prior knowledge on similar calibration procedures. All subjects were supplied with short one-page user manuals on how to use the calibration interfaces outlined in Section \ref{section:software_toolchain_for_flexible_teaming}. After reading these manuals, four users executed both calibration steps. We measured the reading time needed to comprehend the manuals and the time needed for each step (Table \ref{tab:calibration_eval}). All in all, none of the participants exceeded an overall time of ca. 20min between reading the manual and completing calibration. The one-click camera calibration is straightforward and thus did not raise any questions beyond the manual. It was performed correctly by all users at the first attempt. LIDAR calibration is more complex and demands users to understand how to collect beneficial correspondences and when sufficient precision has been reached. These issues resulted in a low number of questions across all users. Still, all of them succeeded at least at the second attempt.
\begin{table}
	\centering
	\begin{tabular}{@{}p{0.12cm}p{2.5cm}rrrr@{}}
		\toprule
		\multicolumn{2}{m{2cm}}{} & P1 & P2 & P3 & P4 \\ 
		\hline
		\multicolumn{2}{@{}l}{Camera} & & & & \\
		\cdashline{1-6}
		 & Reading Time [min] & 0:44 & 0:40 & 0:18 & 0:24 \\ 
		
		& Execution Time [min] & 6:42 & 6:50 & 6:20 & 6:27 \\ 
		
		& \# of questions asked & 0 & 0 & 0 & 0 \\ 
		\hline
		\multicolumn{2}{@{}l}{LIDAR} & & & & \\
		\cdashline{1-6}
		& Reading Time [min] & 1:01 & 3:16 & 1:08 & 2:02 \\ 
		
		& Execution Time [min] & 6:50 & 6:30 & 5:21 & 10:13 \\ 
	
		& \# of attempts & 2 & 1 & 1 & 2 \\ 
		
		& \# of questions asked & 2 & 3 & 1 & 3 \\ 
		\hline
		\multicolumn{2}{c}{Overall Calibration Time [min]} & 15:17 & 17:16 & 13:07 & 19:06 \\
		\bottomrule
	\end{tabular} 
	\caption{User evaluation results}
	\label{tab:calibration_eval}
	\vspace{-1cm}
\end{table}

\textbf{Task Modelling:} Our user evaluation presented in \cite{Riedelbauch18} shows, that non-experts can handle skill-based modelling of precedence graphs for pick\&place tasks intuitively. We could furthermore show, that complex graphs with up to 84 elements can be modelled in less than 10 minutes after gathering some experience with the editor.

\textbf{Online Task-Sharing:} Preliminary prior results from simulated human-robot shared task executions demonstrate, that flexible teaming can accelerate task execution \cite{Riedelbauch19}. The qualitative user evaluation of our human-robot communication app indicate, that users generally accept this mode of interaction despite a need for improvements regarding the user interface and communication timing \cite{Riedelbauch19b}.


\section{CONCLUSION AND FUTURE WORK}\label{section:conclusion_and_future_work}
Teams composed of humans and robots will be an essential part of Factories of the Future. The \textsc{FlexCobot} project investigates flexible teaming with dynamic task allocation and transitions between coexistence, cooperation, and collaboration. We have contributed a toolchain that supports end-users throughout all stages of system operation, from initial setup by calibration via task modelling to dynamic execution. This short paper complements our prior experiments with an initial user evaluation of the calibration step. Combining qualitative previous results with those regarding calibration we suggest, that the overall system may be operated by end-users without expert knowledge on robotics. We have moreover observed from a quantitative point of view, that calibration can me managed in less than 20 minutes, while composing complex task models takes no more than ten minutes. Thus, the span of time between having the system hardware installed and deploying the first shared task can be kept below 30 minutes. These results are of course limited by the application domain implemented in the prototype. We currently support pick\&place operations with simple objects, and an assembly skill, where the robot holds a receiving part, while the human attaches mounting part. Although these skills cover typical applications conceptually, future work may target the extension and re-evaluation of the toolchain in the context of industrial use cases. Furthermore, questions during the calibration user evaluation indicate a need for a more intuitive indicator for LIDAR calibration precision. One may address this issue by providing a traffic light alike feature based on reprojection errors before launching large-scale evaluation of the overall system.

\bibliographystyle{IEEEtran}
\bibliography{library}

\end{document}